%% file: main.tex
\documentclass[11pt]{article}
\usepackage[margin=1in]{geometry}
\PassOptionsToPackage{numbers, compress}{natbib}

\usepackage[utf8]{inputenc} 
\usepackage[T1]{fontenc}    
\usepackage{hyperref}       
\usepackage{url}            
\usepackage{booktabs}       
\usepackage{amsfonts}       
\usepackage{nicefrac}       
\usepackage{microtype}      
\usepackage{xcolor}         
\usepackage{enumitem}
\usepackage{natbib}

\usepackage{graphicx}
\usepackage{url}
\usepackage[noend]{algorithm2e}
\usepackage{amsmath}
\usepackage{booktabs}
\usepackage{wrapfig}
\usepackage{float}
\usepackage{enumitem}

\makeatletter
\renewcommand*{\@fnsymbol}[1]{\ensuremath{\ifcase#1\or \dagger\or \ddagger
\else\@ctrerr\fi}}
\makeatother

\newcommand\blfootnote[1]{%
  \begingroup
  \renewcommand\thefootnote{}\footnote{#1}%
  \addtocounter{footnote}{-1}%
  \endgroup
}

\title{Federated LoRA with Sparse Communication}
\author{
Kevin Kuo~\thanks{Computer Science Department, Carnegie Mellon University, Pittsburgh, Pennsylvania, USA}
\and Arian Raje~\footnotemark[1]
\and Kousik Rajesh~\thanks{Machine Learning Department, Carnegie Mellon University, Pittsburgh, Pennsylvania, USA}
\and Virginia Smith~\footnotemark[2]
}

\newcommand{\ours}{\textbf{\textsc{FLASC}}}
\newcommand{\sparseadapter}{\textbf{\textsc{SparseAdapter}}}
\newcommand{\adapterlth}{\textbf{\textsc{Adapter LTH}}}
\newcommand{\lora}{{\textsc{LoRA}}}
\newcommand{\fedadam}{{\textsc{FedAdam}}}
\newcommand{\ffalora}{\textbf{\textsc{FFA-LoRA}}}
\newcommand{\flselect}{\textbf{\textsc{Federated Select}}}
\newcommand{\hetlora}{\textbf{\textsc{Heterogeneous LoRA}}}
\newcommand{\compeft}{\textbf{\textsc{ComPEFT}}}

\begin{document}

\maketitle
\begin{abstract}
\normalsize
\input{src/abstract}\blfootnote{Preprint. Under review.}
\end{abstract}

\section{Introduction}
\input{src/introduction}

\section{Related Work}\label{sec:relwork}
\input{src/relwork}

\section{Empirical Results}
\input{src/results}

\section{Conclusion and Future Work}
\input{src/conclusion}

\section{Acknowledgements}
We thank Steven Kolawole and Don Dennis for their helpful comments.

\pagebreak
\bibliography{main.bib}


\appendix


\input{src/appendix}


\end{document}

%% file: src/abstract.tex
Low-rank adaptation (LoRA) is a natural method for finetuning in communication-constrained machine learning settings such as cross-device federated learning. Prior work that has studied LoRA in the context of federated learning has focused on improving LoRA's robustness to  heterogeneity and privacy. In this work, we instead consider techniques for further improving communication-efficiency in federated LoRA. Unfortunately, we show that centralized ML methods that improve the efficiency of LoRA through unstructured pruning do not transfer well to federated settings. We instead study a simple approach, \textbf{FLASC}, that applies sparsity to LoRA during communication while allowing clients to locally fine-tune the entire LoRA module. Across four common federated learning tasks, we demonstrate that this method matches the performance of dense LoRA with up to $10\times$ less communication. Additionally, despite being designed primarily to target communication, we find that this approach  has benefits in terms of heterogeneity and privacy relative to existing approaches tailored to these specific concerns. Overall, our work highlights the importance of considering system-specific constraints when developing communication-efficient finetuning approaches, and serves as a simple and competitive baseline for future work in federated finetuning.

%% file: src/introduction.tex
As pretrained models continue to advance state-of-the-art performance in a variety of domains, it is critical to develop methods for efficiently finetuning models in low-resource settings. In this work we consider the cross-device federated learning (FL) setting which seeks to train models across a network of heterogeneous edge devices~\citep{mcmahan2017communication}. A major bottleneck in FL is the cost of \textit{communicating model updates} from the client to the server, which can make finetuning large models prohibitive~\citep{konecny2017federated}. 

Recently, adapter methods have emerged as an effective way to reduce costs in both centralized and federated settings~\citep{houlsby2019parameter,hu2021lora,zhang2023fedyolo}. In this work, we focus on \textit{low-rank adaptation} (\lora), a popular method which injects trainable low-rank adapters into a model and freezes the pretrained backbone. Although \lora~is able to match the performance of full (backbone) finetuning with much (e.g. $100\times$) fewer parameters, communicating these parameters can still be expensive if the adapter is specified with a large raw number of parameters or the network is slow. To further reduce the parameters in \lora, recent work in the centralized setting has proposed using \textit{unstructured pruning}, which zeros and freezes a large fraction of the adapter weights~\citep{wu2022pruning, he2022sparseadapter}. However, as we show, these ``adapter pruning'' schemes transfer poorly to FL applications because they \textit{a)} have limited utility due to freezing weights~\citep{raihan2020sparse} and \textit{b)} are unable to handle asymmetric communication speeds~\citep{konecny2017federated,ro2022scaling}.

Beyond limited communication, data and/or systems heterogeneity and privacy are commonly studied issues in FL which can harm model training. Since pretraining has been shown to help mitigate these issues to some extent~\citep{nguyen2022begin,li2021large}, it is natural to ask whether \lora~is effective in handling these concerns. Unfortunately, recent works suggest that challenges with heterogeneity and privacy in federated settings can in fact be exacerbated by using \lora~\citep{babakniya2023slora,sun2024improving}. 

In this work, we present \ours~(\textbf{F}ederated \textbf{L}ow-Rank \textbf{A}daptation with \textbf{S}parse \textbf{C}ommunication)---a method that improves the communication-efficiency of federated \lora~while also having favorable performance in terms of other common challenges such as heterogeneity and privacy. As its name suggests, \ours~applies \textit{sparse communication to} \lora.  In \ours, clients download a sparse \lora~module; unlike pruning-based approaches, we allow clients to finetune all the LoRA parameters, rather than keeping the zeroed weights frozen after applying sparsity. During upload, the dense local update is sparsified using a potentially different sparsity pattern than download. This simple adjustment greatly improves utility and has little efficiency downsides. Furthermore, our method allows for separate configuration of download and upload sparsity, making it well-suited for FL settings constrained by upload bandwidth. 

Overall our work makes the following contributions:
\begin{enumerate}[leftmargin=23pt]
    \item 
    We study the use of unstructured sparsity in \lora~to enable communication-efficient federated finetuning. Our work identifies a key limitation in existing methods: freezing weights dramatically limits model utility, yet provides little practical efficiency gains.
    \item We propose \ours, a simple method that applies sparsity to \lora~by only communicating Top-K magnitude entries. Our method can reduce communication costs up to 10$\times$ while matching the performance of \lora~on several FL image and text tasks.
    \item We conduct extensive experiments which show that both \lora~and our method are robust to heterogeneous and private cross-device FL. In such settings, our method reduces the communication of \lora~by up to $16\times$ while outperforming other sparsity and freezing-based methods designed for these concerns. 
\end{enumerate}

%% file: src/relwork.tex
\vspace{-.1in}
\textbf{Communication-efficient federated learning.} Communication, particularly on upload, is a key bottleneck in federated settings due to slow network bandwidth. While large pretrained models can significantly boost utility in federated settings~\citep{radford2018improving,nguyen2022begin}, these models present new challenges with communication and finetuning over the edge. Many types of methods have been explored to reduce FL communication costs, including quantization~\citep{reisizadeh2020fedpaq,ozkara2021quped}, sparsity~\citep{caldas2018expanding,horvath2021fjord,bibikar2022federated,stripelis2022federated,isik2022sparse}, and parameter-efficient finetuning (PEFT)~\citep{chen2023federated,babakniya2023slora}. PEFT methods are surprisingly effective in FL; for example, \lora~can train an adapter over $100\times$ smaller than the original model~\citep{hu2021lora} while more complex FL methods degrade noticeably when compressing the backbone beyond $10\times$~\citep{qiu2021zerofl,babakniya2023revisiting,ro2022scaling}.

\textbf{Parameter-efficient finetuning (PEFT)} reduces the cost of finetuning by training a small number of parameters and freezing the rest of the model~\citep{ding2022delta}. In this work, we focus on \textit{low-rank adaptation} (\lora), a popular reparameterization-based method which has two advantages: First, \lora~can be merged with the backbone after training, which removes its additional inference costs~\citep{houlsby2019parameter,hu2021lora}. Second, prior work has shown that \lora~achieves better efficiency-utility trade-offs than other PEFT methods based on pruning and backbone finetuning~\citep{guo2021parameter,zaken2022bitfit,sung2021training,gong2022finding}. 

\textbf{Efficient \lora.} Recent works in the centralized setting that improve \lora's efficiency consider unstructured sparsity~\citep{wu2022pruning,he2022sparseadapter}, structured sparsity~\citep{ding2023sparse,liu2024sparsely}, quantization~\citep{xu2023qa,dettmers2024qlora}, or flexibly adjusting the rank~\citep{zhang2023increlora,zhang2022adaptive}. Beyond directly modifying the \lora~parameters, \lora~can also be used to efficiently prune or update the backbone parameters~\citep{zhao2023cpet,zhang2023pruning,zhao2024apt}. Later, we show that centralized ``sparse \lora'' baselines are ineffective in FL settings but can be made extremely efficient by only targeting communication.

\textbf{Federated LoRA.} In the federated setting, many works have observed that \lora~substantially reduces the communication cost of FL finetuning~\citep{sun2022conquering,malaviya2023reducing,zhang2023fedpetuning,nguyen2024flora}. Follow-ups to these works have raised potential challenges with extending \lora~to FL, such as data heterogeneity~\citep{kim2023client, yi2023fedlora,lu2024hyperflora,jiang2024personalized,babakniya2023slora}, systems heterogeneity~\citep{cho2023heterogeneous,bai2024federated}, differential privacy~\citep{sun2024improving}, and multi-modal data~\citep{chen2024feddat,ping2024fl}. Our work is most similar to \compeft, which considers one-shot merging of compressed \lora~adapters uploaded from multiple sources~\citep{yadav2023compeft}. In contrast, we study methods for reducing both upload and download communication over multiple rounds of FL training. Overall, we find that unstructured sparsity can significantly improve the communication of \lora~without significantly impacting its robustness to the concerns listed above.

\begin{figure}[t]
    \centering
    \includegraphics[width=13.5cm]{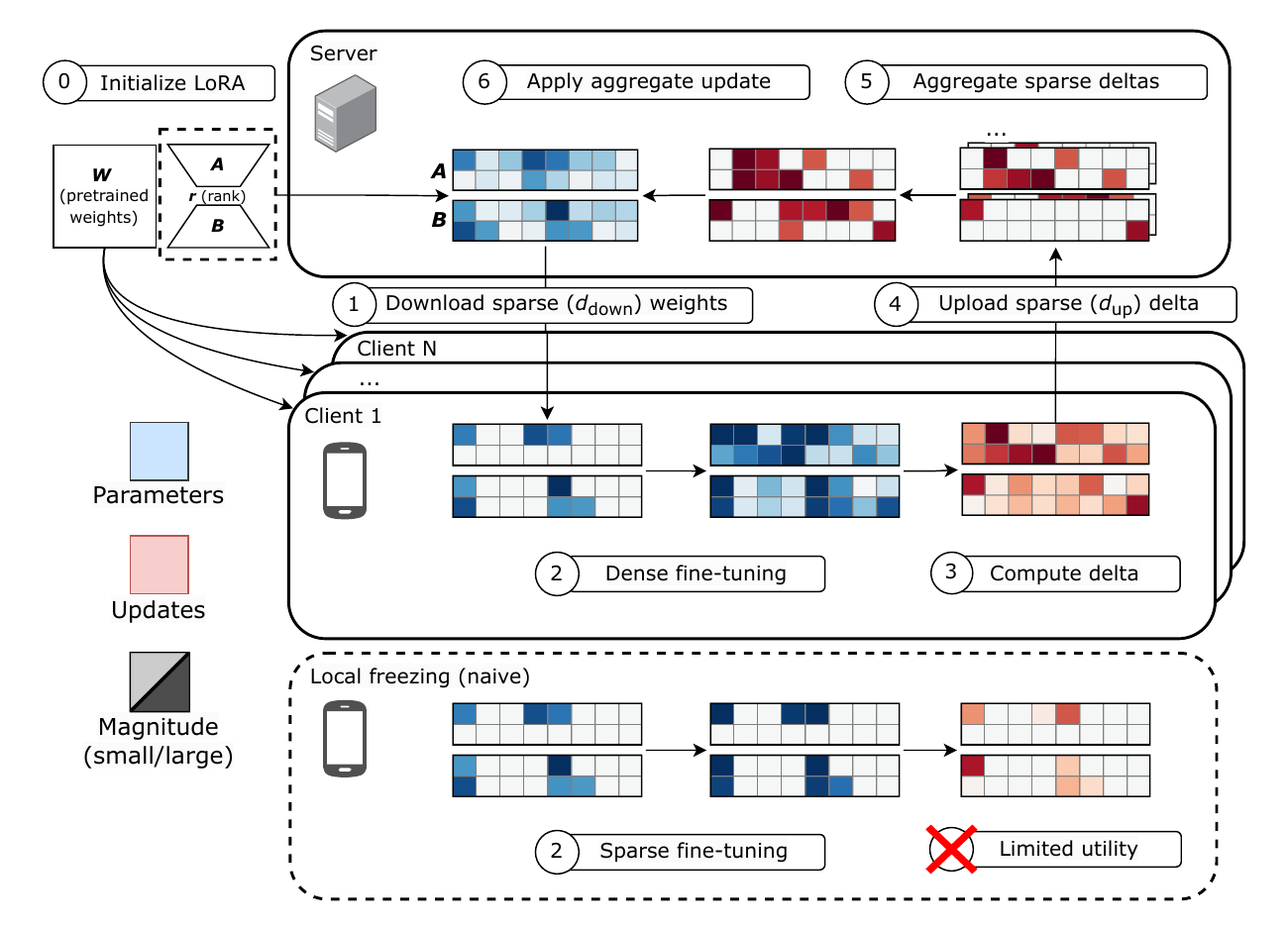}
    \caption{A step-by-step overview of \ours. Step 0 is executed prior to FL training, while training repeats steps 1-6. \color{blue}Blue\color{black}/\color{red}red \color{black} squares indicate the magnitude of \color{blue}weights\color{black}/\color{red}updates \color{black}respectively. \textbf{Darker} squares indicate a \textbf{larger magnitude}, which is the ranking criterion $(\ell_1)$ used for sparsity. }
    \label{fig:overview}
\end{figure}
\section{Federated Low-Rank Adaptation with Sparse Communication}
\label{sec:method}
\vspace{-.1in}
Low rank adaptation (\lora)~is a reparameterization-based PEFT method that updates a weight matrix $W \in \mathbb{R}^{d \times k}$ in a low-rank subspace~\cite{hu2021lora}. \lora~freezes $W$ and defines the update $\Delta W \in \mathbb{R}^{d \times k}$ as a product $BA$ where $B \in \mathbb{R}^{d\times r}$ and $A\in\mathbb{R}^{r\times k}$ are newly inserted trainable parameters. To apply \lora~to FL, we can simply treat the adapter weights $(A,B)$ as the global trainable weights for a federated optimization method such as \fedadam~(see Appendix~\ref{appendix:methods}). 

Assuming that all clients have a copy of the pretrained model, only the \lora~parameters need to be communicated and updated at each round. However, communicating the \lora~parameters may still be costly, particularly in communication-constrained federated networks. We thus consider \lora~as a naive baseline for federated finetuning and explore how to further reduce its message size using sparsity.

\textbf{Pruning} methods are a natural choice to improve the communication efficiency of \lora. To apply these methods to \lora, we can prune weights in the adapters $(A,B)$, setting them to zero and freezing them for the rest of training. A key design choice when applying pruning is the \textit{granularity} of sparsity to apply. \textit{Structured} (group-level) sparsity has the advantage of reducing computation without needing a specialized setup, while \textit{unstructured} (parameter-level) sparsity tends to achieve higher utility~\citep{liu2018rethinking, siswanto2021reconciling}. As we explain next, pruning \lora~adapters generally yields limited computational benefits, making ``unstructured sparse \lora'' uniquely suited for concerns of communication.

To our knowledge, there are two existing works that apply unstructured sparsity to \lora: \adapterlth~and \sparseadapter~\citep{wu2022pruning,he2022sparseadapter}. These two methods correspond to two canonical algorithms for training sparse models: \textit{Iterative magnitude pruning} gradually prunes the model while re-training the remaining weights~\citep{renda2019comparing}. In contrast, \textit{pruning-at-initialization} performs one-shot pruning followed by a single sparse re-training stage~\citep{lee2018snip,tanaka2020pruning,wang2019picking}. Unfortunately, naive application of pruning methods on top of \lora~may not yield significant efficiency gains, even in centralized settings. In particular, a sparse \lora~adapter can be stored more efficiently, but otherwise has marginal computational benefits. This is because:

\begin{itemize}[leftmargin=15pt]
    \item \textbf{The compute and memory costs of adapters are small} compared to the costs of the backbone~\citep{kim2024memory}. Additionally, reparameterization-based PEFT modules such as \lora~can be merged with the backbone once training is complete~\citep{luo2023towards}. This eliminates adapter inference costs and makes it less important to produce an extremely sparse adapter.
    \item \textbf{Unstructured sparsity often requires specialized hardware and software} to accelerate computation. Without the proper setup, sparse training and inference are no more efficient than that of a dense counterpart~\citep{muralidharan2023uniform}. 
\end{itemize} 
Despite these limitations, combining unstructured sparsity with LoRA is particularly effective for handling issues of communication in FL. Additionally, we show that it is important to carefully incorporate sparsity with \lora~to see benefits in terms of communication. Our proposed method, \ours~(Figure~\ref{fig:overview}), has three key features which enable it to perform well in FL settings:
\begin{enumerate}[leftmargin=15pt]
    \item \textbf{Local finetuning uses dense gradients.}  As mentioned above, pruning methods freeze the pruned weights in order to reduce computation. Unfortunately, freezing can lead to lower utility compared to dense training, which is often considered an ideal baseline in the pruning literature~\citep{dettmers2019sparse,sung2021training}. However, because of the marginal compute differences between dense and sparse finetuning of \lora, we opt to use dense finetuning for its superior utility.
    \item \textbf{Upload and download sparsity are applied independently.} When global parameters are frozen, clients have an identical sparse structure on download and upload. In contrast, \ours~masks dense local updates before uploading. This improves utility and conveniently enables us to target upload and download at varying sparsities, making \ours~a natural choice for practical FL settings where upload speeds are typically much slower than download speeds (up to $8\times$)~\citep{konecny2017federated,lai2022fedscale}.
    \item \textbf{The download mask can change across rounds.} Adaptively (un)freezing parameters during training improves performance over methods which fix a sparse structure throughout~\citep{qiu2021zerofl,bibikar2022federated,babakniya2023revisiting}. \ours~applies this idea to download communication by having the server maintain a dense set of weights and temporarily apply a Top-K magnitude mask during download.
\end{enumerate}

Pseudocode is provided in Algorithm~\ref{alg:pseudocode}. $P$ refers to a flattened and concatenated vector of LoRA weights $\{A_l, B_l\}_{l=1}^L$ where $L$ is the number of layers LoRA is applied to. We apply \textit{global} sparsity i.e. retain the Top-K magnitude entries of $P$. An alternative approach is to uniformly sparsify each layer $(A_l, B_l)$ in a \textit{layer-wise} way before concatenation, but we found that global sparsity tended to perform better.

 \input{src/pscode}

%% file: src/pscode.tex
\RestyleAlgo{boxruled}
\LinesNumbered
\IncMargin{2em}
\SetAlCapHSkip{1em}
\begin{algorithm}[H]
\caption{PyTorch-like pseudocode for \ours}\label{alg:pseudocode}
Require: $d_\text{down}, d_\text{up}$ (download and upload density), $r$ (\lora~rank) \\
$P \gets$ Initialize \lora~($\text{rank}=r$) parameters\\
$\texttt{optim} \gets $ \texttt{torch.nn.optim.Adam(params=}$P$\texttt{)} \\
\For{$r=1,...,R$}{
    $M_\text{down} \gets $ mask of top $d_\text{down}$ fraction entries of $P$ by magnitude \\
    Sample clients $c_1, ..., c_n$ uniformly at random without replacement\\
    \For{$i = 1, ..., n$ in parallel}{
        $P_{i} = P \odot M_\text{down}$ \hspace{4.92cm} \color{red}\# sparse download\color{black}\\
        $P_{i}' \gets$ update $P_i$ with 1 SGD epoch on data of $c_i$ \hspace{0.5cm} \color{red}\# fine-tuning all entries of $P_i$ \color{black}\\
        $\Delta P_{i} \gets P_{i} - P_{i}'$\\
        $M_{\text{up}, i} \gets $ mask of top $d_\text{up}$ fraction entries of $\Delta P_{i}$ by magnitude \\
        $\Delta P_{i} \gets \Delta P_{i} \odot M_{\text{up}, i}$ \hspace{4.32cm} \color{red}\# sparse upload\color{black}\\
    }
    \texttt{optim.grad} $\gets \frac{1}{n}\sum_{i=1}^n \Delta P_i $ \hspace{3.65cm} \# set Adam pseudo-gradient \\
    \texttt{optim.step()} \hspace{5.65cm} \# update $P$ using Adam
}
\end{algorithm}
\DecMargin{1em}

%% file: src/results.tex
In the following sections, we test \ours~in a variety of FL settings and show that it is effective at handling concerns of communication efficiency (\ref{sec:results_comm}), data heterogeneity (\ref{sec:results_data}), systems heterogeneity (\ref{sec:results_sys}), and privacy (\ref{sec:results_dp}). Generally, we find that when \lora~itself performs well in light of these concerns, it enables \ours~to significantly reduce communication while retaining high utility. 

We present experiments on four datasets: CIFAR10, 20NewsGroups, Reddit, and FLAIR \citep{krizhevsky2009learning,lang1995newsweeder,caldas2018leaf,song2022flair}. CIFAR10 and FLAIR are image datasets; the images are resized to $224\times 224$ to match ImageNet, the pretraining dataset for the ViT model architecture. During training, we apply standard data augmentation methods (random crops and flips). 20NewsGroups and Reddit are text datasets; each example (a news headline or Reddit comment) is preprocessed using the GPT2 tokenizer into a sequence with length 128 or 25 respectively. We partition CIFAR10 and 20NewsGroups using a synthetic Dirichlet distribution over the set of labels~\citep{hsu2019measuring}. Reddit and FLAIR are obtained from social media sites (Reddit and Flickr) and are naturally partitioned by user. 

\begin{table}[h]
    \centering
    \begin{tabular}{lcccccc}
        \toprule[\heavyrulewidth]
        \textbf{Dataset} & Task & Partition & \#Clients & \#Examples & \#Classes\\
        \midrule
        CIFAR10 & Image Classification & Dirichlet & 500 & 50K & 10 \\
        20NewsGroups &  Sequence Classification  & Dirichlet & 350 & 20K & 20 \\
        Reddit & Next Token Prediction & Natural & 32K & 1.1M & 50257 \\
        FLAIR & Object Detection & Natural & 41K & 345K & 17 (coarse) \\
        \bottomrule[\heavyrulewidth]
    \end{tabular}
    \vspace{0.05cm}
    \caption{Training partition statistics of the datasets used in the experiments.}
    \label{tab:datasets}
\end{table}

\begin{figure}
    \centering
    \includegraphics[width=\textwidth]{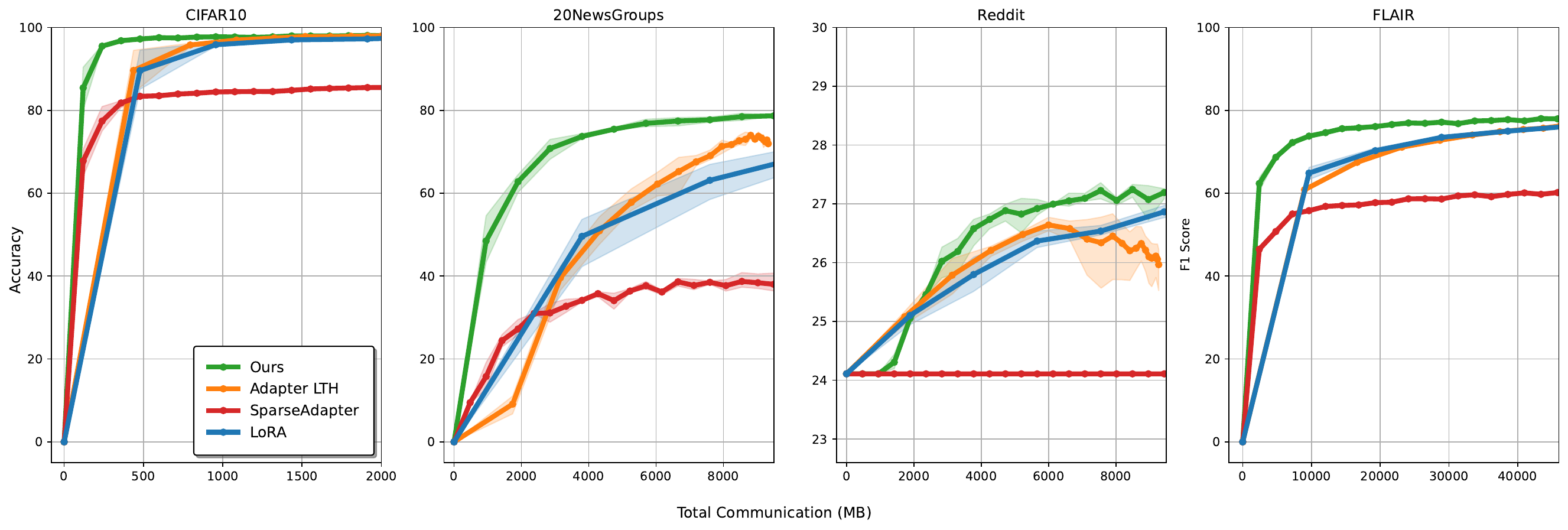}
    \vspace{-.4cm}
    \caption{We compare utility ($\uparrow$) vs. total communication when augmenting LoRA (rank $r=16$) with sparsity. Out of all four methods, \ours~reaches the highest utility with the least communication. In contrast, \adapterlth~is inefficient early in training and \sparseadapter~fails to match the utility of \lora. Shaded bands show the min/mean/max utility over 3 random seeds.}
    \label{fig:methods}
    \vspace{-.2cm}
\end{figure}

We used VIT-B-16 (85M params) and GPT2-Small (124M params) as the backbone for image and text tasks respectively~\citep{dosovitskiy2020image,radford2019language}. For all datasets, we use a local batch size of 16. For FLAIR, we sample 200 clients per round, finetune for 2 local epochs, and communicate for up to 5000 rounds. For the three other datasets, we sample 10 clients each round, finetune for 1 local epoch, and communicate for up to 200 rounds. More details on hyperparameters are in Appendix~\ref{appendix:hps}.

\subsection{Communication Efficiency}\label{sec:results_comm}
First, we measure the communication that each method uses during a single training run. In Figure~\ref{fig:methods}, we show the utility vs. communication of \lora~and compare three methods for reducing its communication: \adapterlth, \sparseadapter, and \ours~(described in Section~\ref{sec:method}). We set \adapterlth~to keep 0.98$\times$ (prune away 2\%) of the remaining weights every 25 rounds for FLAIR and every round for the other 3 datasets. We use a density of $0.25\times$ for \sparseadapter~and \ours~(upload and download). \ours~can use a much lower density on some datasets, but we use $0.25\times$ density across all datasets to simplify the comparison.

\begin{wrapfigure}{h}{0.5\textwidth} 
 \vspace{-.5cm}
\includegraphics[width=0.5\textwidth]{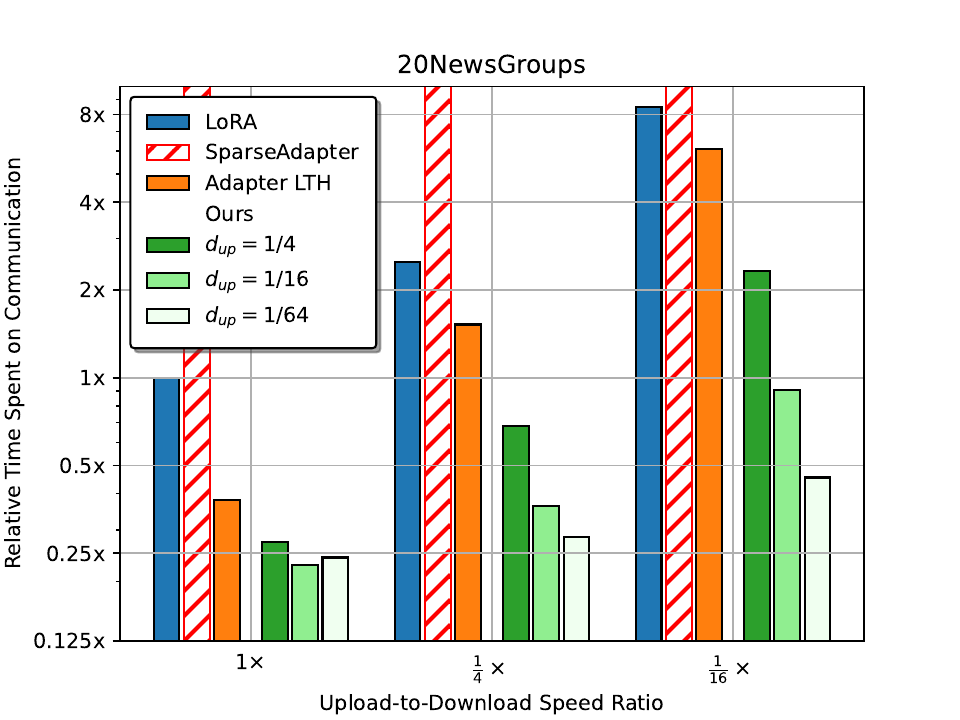}
\vspace{-.2in}
\caption{We measure the communication time ($\downarrow$) needed to reach 70\% accuracy on 20NewsGroups. Beyond efficiency in terms of total communication $(1\times)$, \ours~(green) is robust to extremely slow upload speed $(16\times)$ by making upload more sparse than download. Hatched bars indicate that \sparseadapter~failed to reach 70\% accuracy.}
\vspace{-.1in}
\label{fig:time}
\end{wrapfigure}

Across the four tasks, \ours~is able to match the performance of \lora~while using $3$--$10\times$ less communication. In contrast, the other two methods cannot reliably match the performance of~\lora. Despite a relatively large density of $0.25\times$, \sparseadapter~always fails to match the performance of \lora. \adapterlth~has limited benefits due to its iterative nature; in early rounds, it uses a similar amount of communication as \lora, while in later rounds, the adapters are too sparse to continue training and performance plateaus or degrades.

In Figure~\ref{fig:time}, we measure the communication time that each method needs to reach 70\% accuracy on 20NewsGroups. We consider three settings where the upload bandwidth is $\{1,1/4,1/16\}$ times that of the download bandwidth. To simplify the analysis, we assume ideal noiseless channels where communication time is equal to the size of the \lora~update divided by a fixed bandwidth. Because bandwidth capabilities vary across systems, we instead show the communication time of each method as a \textit{ratio of the time spent} relative to dense \lora~($1\times$). First, we explore the same sparsity methods as Figure~\ref{fig:methods}: \sparseadapter~$(d=\frac{1}{4})$~fails to reach 70\% accuracy, while \adapterlth~($d=0.98$)~has moderate efficiency gains. Meanwhile, \ours~$(d_{\text{down}}=d_{\text{up}}=\frac{1}{4})$ is the most robust and spends $\sim 4\times$ less time on communication than \lora~across all 3 settings. However, if bandwidth conditions are known, \ours~can handle slow uploads even more effectively by further reducing the upload density. In particular, uploads as sparse as $d_\text{up}=\frac{1}{64}$ can achieve 70\% accuracy on 20NewsGroups with $\sim 16\times$ faster communication than \lora.

\subsection{Sparsity without Freezing} \label{sec:results_freeze}
Although weight freezing usually harms utility, simple freezing methods can still serve as competitive baselines against complex pruning methods which \textit{adaptively} freeze and unfreeze weights. Naively applying such methods to FL can significantly degrade utility, while freezing can avoid such failure modes~\citep{babakniya2023revisiting}. Therefore, a key question in the design of \ours~is how the server and clients should apply freezing alongside sparsity. 

In the next experiment, we study two representative baselines for training sparse \lora~which differ in the way they freeze parameters. \sparseadapter~selects a global subset of \lora~weights after one FL round and freezes all other weights for the rest of training. \flselect~incorporates the idea of server-side adaptivity: clients still download (select) a global sparse model and only finetune these sparse parameters, but the downloaded parameters can change across rounds~\citep{charles2022federated}. To~summarize, both methods freeze weights on the \textit{client level}, which naturally fixes the upload sparsity structure to match that of download. Additionally, \sparseadapter~freezes weights on the \textit{server level}, which fixes the same download structure across all rounds. Finally, \ours~does not freeze at all; we compute dense updates and apply sparsity only during communication.

\begin{wrapfigure}{h}{0.5\textwidth}
\includegraphics[width=0.5\textwidth]{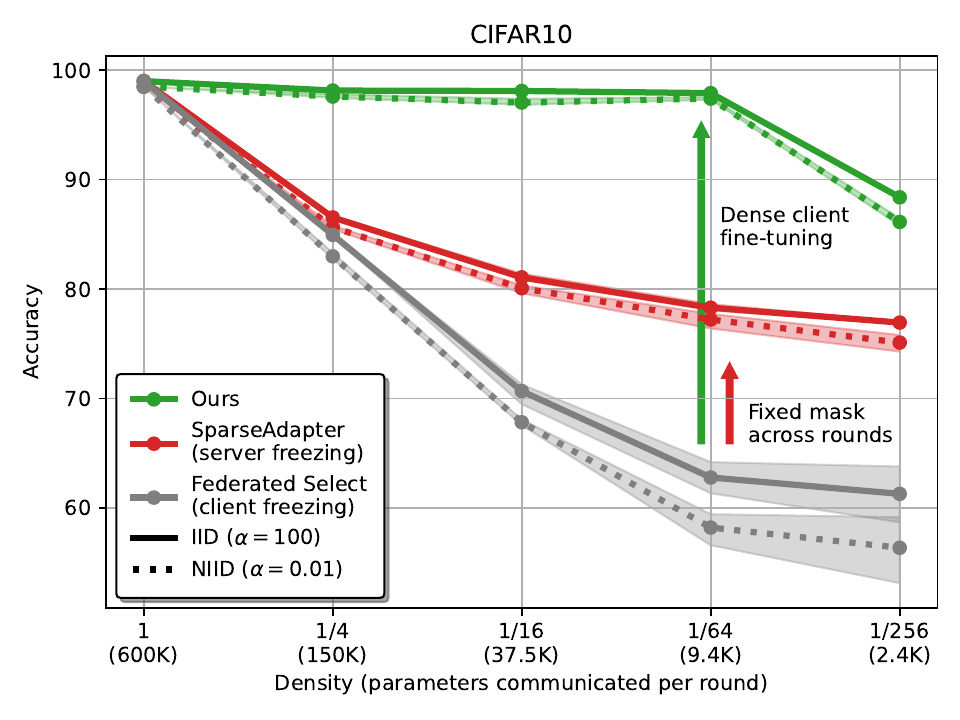}
\caption{We compare the accuracy $(\uparrow)$ of \ours~to two ways of freezing weights while training an unstructured sparse $\lora~(r=16)$ module with \fedadam.}
\label{fig:iid_freezing}
\end{wrapfigure}

In Figure \ref{fig:iid_freezing}, we find that \flselect~performs extremely poorly, as this method was not designed for \lora~nor unstructured sparsity. \sparseadapter~works better across all density values despite employing a relatively simpler method. Similar works have also found that global freezing can perform reasonably well in FL settings~\citep{babakniya2023revisiting}. Finally, \ours~greatly improves utility over both methods by only considering sparse communication without any freezing at all. As discussed in Section~\ref{sec:method}, freezing \lora~parameters has relatively small compute savings, suggesting that we should try to leverage the utility of dense local updates and then reduce commmunication afterwards. Surprisingly, we find that dense local updates can be sparsified to an even higher degree than what is achievable with sparse finetuning, despite the additional ``density'' introduced by dense finetuning.

\begin{figure}
\centering
    \includegraphics[width=1.0\textwidth]{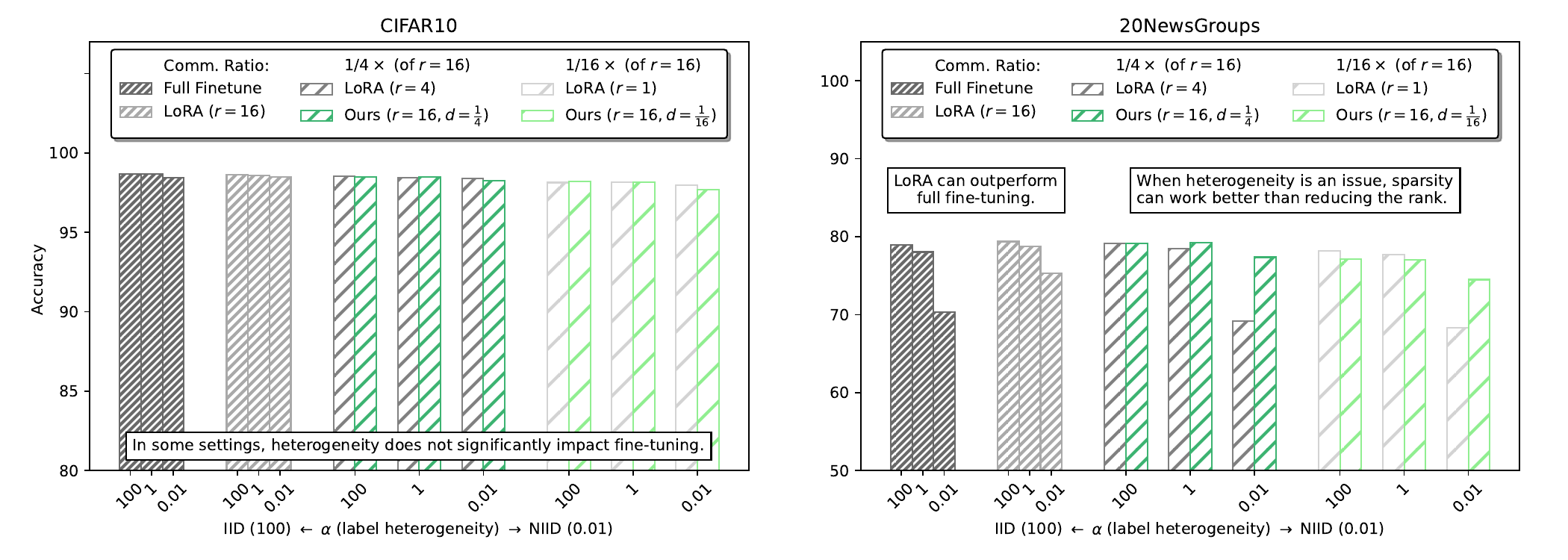}
    \vspace{-0.4cm}
    \caption{We show accuracy ($\uparrow$) in settings with varying label heterogeneity. We reduce communication using a) lower \lora~rank or b) sparsity with \ours. Bars are grouped by communication cost and ordered by increasing heterogeneity (decreasing $\alpha$). We find that tuning the rank is important; \ours~with $r=16$ can outperform full finetuning and smaller ranks with similar communication.}
    \label{fig:iid}
    \vspace{-0.2cm}
\end{figure}

\subsection{Data Heterogeneity} \label{sec:results_data}

Statistical heterogeneity is a commonly studied issue in FL due to the its negative effect on the utility and convergence of FL optimization methods~\citep{li2020federated}. Following standard practice in FL literature, we partition a centralized dataset by drawing samples from a Dirichlet distribution to determine the clients' label distributions~\citep{hsu2019measuring}. In Figure~\ref{fig:iid}, we test three values for the heterogeneity parameter $\alpha$. At $\alpha=100$, clients have an approximately uniform number of examples per label, while at $\alpha=0.01$, over $90\%$ of each client's examples belong to a single label. 

The first two groups of bars on the left compare full finetuning to \lora~($r=16$). The paired bars in the middle compare two methods of reducing the communication of \lora~by $\sim 4\times$: either lower the rank $(r=4)$ of \lora~or apply sparsity $(d_\text{up} = d_\text{down} = 1/4)$ using $\ours$. Finally, the group of bars on the right show a $\sim 16\times$ reduction in communication.

On CIFAR10 (left), we find that label heterogeneity has a small negative effect but no significant impact across methods. Since \lora~assumes the use of a pretrained model, it is reasonable to encounter results where the pretrained initialization is robust to heterogeneous data~\citep{zhang2023fedpetuning}. On 20NewsGroups (right), heterogeneity has a more significant effect. In such cases, we find that it is important to tune the rank. In particular, when $\alpha=0.01$ on 20NewsGroups, $r=16$ achieves $\sim 75\%$ accuracy, outperforming full finetuning as well as smaller \lora~ranks $r\in\{1,4\}$ which all achieve $\sim 70\%$ accuracy. By applying sparsity on top of \lora~($r=16$), \ours~can outperform \lora~with a smaller rank with roughly equal communication cost, highlighting the benefits of sparsity alongside tuning the rank of \lora.

\subsection{Systems Heterogeneity}\label{sec:results_sys}
\begin{wrapfigure}{h}{0.5\textwidth}
\vspace{-.4in}
\includegraphics[width=0.5\textwidth]{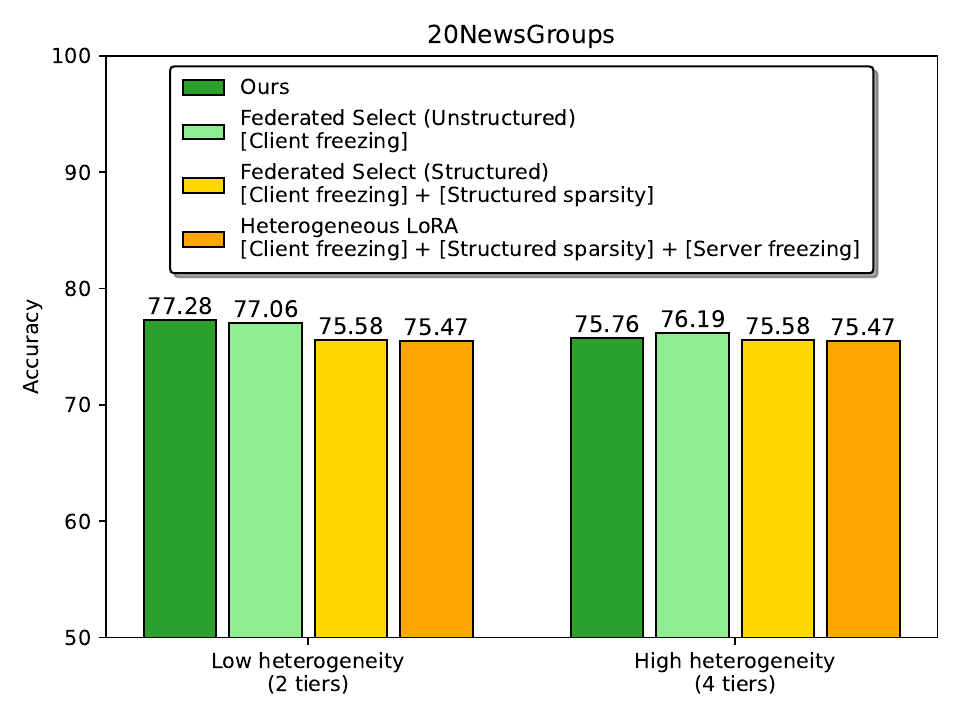}
\includegraphics[width=0.5\textwidth]{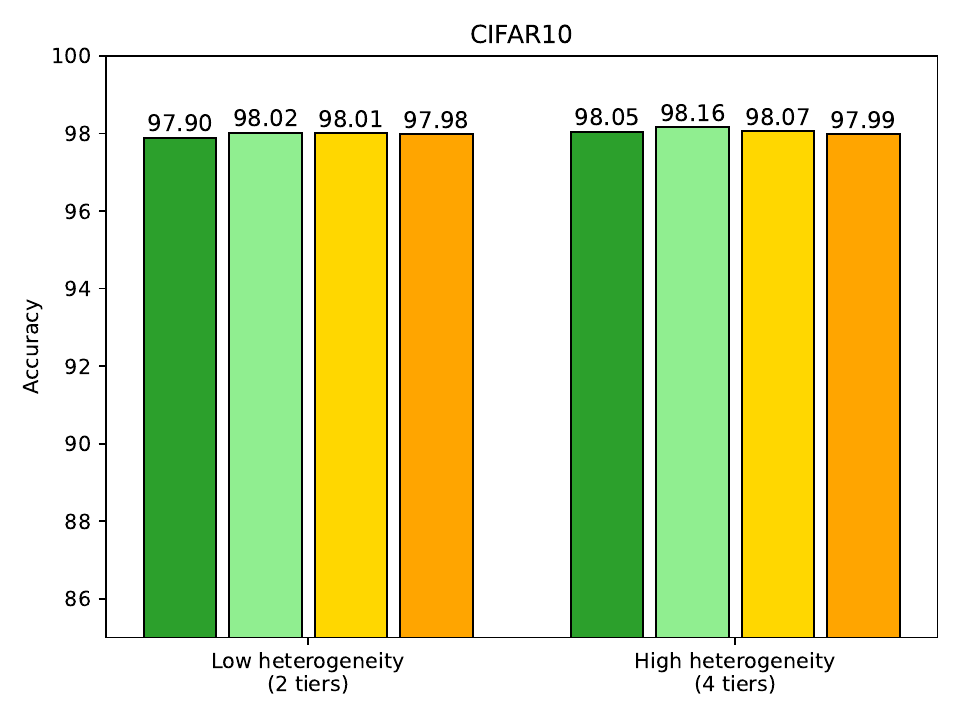}
\vspace{-.2in}
\caption{We compare accuracy $(\uparrow)$ in settings with systems heterogeneity. We find that all methods perform reasonably well, despite having significant differences in freezing and sparsity granularity.}
\vspace{-.3in}
\label{fig:sys}
\end{wrapfigure}
In addition to data heterogeneity, FL settings also face issues of systems heterogeneity. In such settings, clients with low computational resources can slow down or harm the overall utility of training. In recent work, ~\citet{cho2023heterogeneous} develop \hetlora, which aims to address systems heterogeneity by training \lora~modules with different ranks across clients. During training, client $c$ with assigned rank $r_c$ will download the uppermost $r_c$ rows of $A$ and leftmost $r_c$ columns of $B$ from the global \lora~weights (with rank $r_s$), then use the weights to initialize a local \lora~module with rank $r_c$. \ours~can be applied with roughly equal communication cost by finetuning a rank $r_s$ module and sparsifying communication to a density of $d = r_c / r_s$. 

For the experiments, we assign each client $c$ to a budget $b_c \in \{1,2,...,b_s\}$ uniformly at random, where $b_s$ is the number of budget tiers. We consider two heterogeneity settings, low ($b_s=2$) and high ($b_s=4$). For \hetlora, we assign clients a local rank $r_c = 4^{b_c}$, while for \ours, we assign clients a density of $(1/4)^{(b_s - b_c)}$. For both methods, the server initializes a \lora~adapter with rank $r=4^{b_s}$.

\hetlora~bears many similarities with pruning methods. While \hetlora~defines a unique structured mask for each budget, it resembles \sparseadapter~in that these masks do not change over the course of FL. To further investigate the utility of freezing-based methods in systems-heterogeneous settings, we also evaluate \flselect~\cite{charles2022federated}, an intermediate version of \ours~and~\hetlora~that allows the server to adjust the structured sparsity mask associated with each client budget tier.


In Figure~\ref{fig:sys}, we find that \ours~is competitive with \hetlora. Surprsisingly, we also find that \flselect~performs just as well as these two methods, despite the results in Section~\ref{sec:results_freeze} which show that client freezing performs extremely poorly. Our results suggest that freezing is less of an issue when considering systems heterogeneity. Since systems heterogeneity assumes multiple tiers of client resources, there are subsets of clients who can finetune a relatively dense subset of the adapter. Therefore, the gains from adaptively unfreezing weights can be limited. Overall, we find that \ours~is well suited for issues of systems heterogeneity with no additional complexity in terms of configuring the method.

\begin{figure}
    \centering    \includegraphics[width=0.62\textwidth]{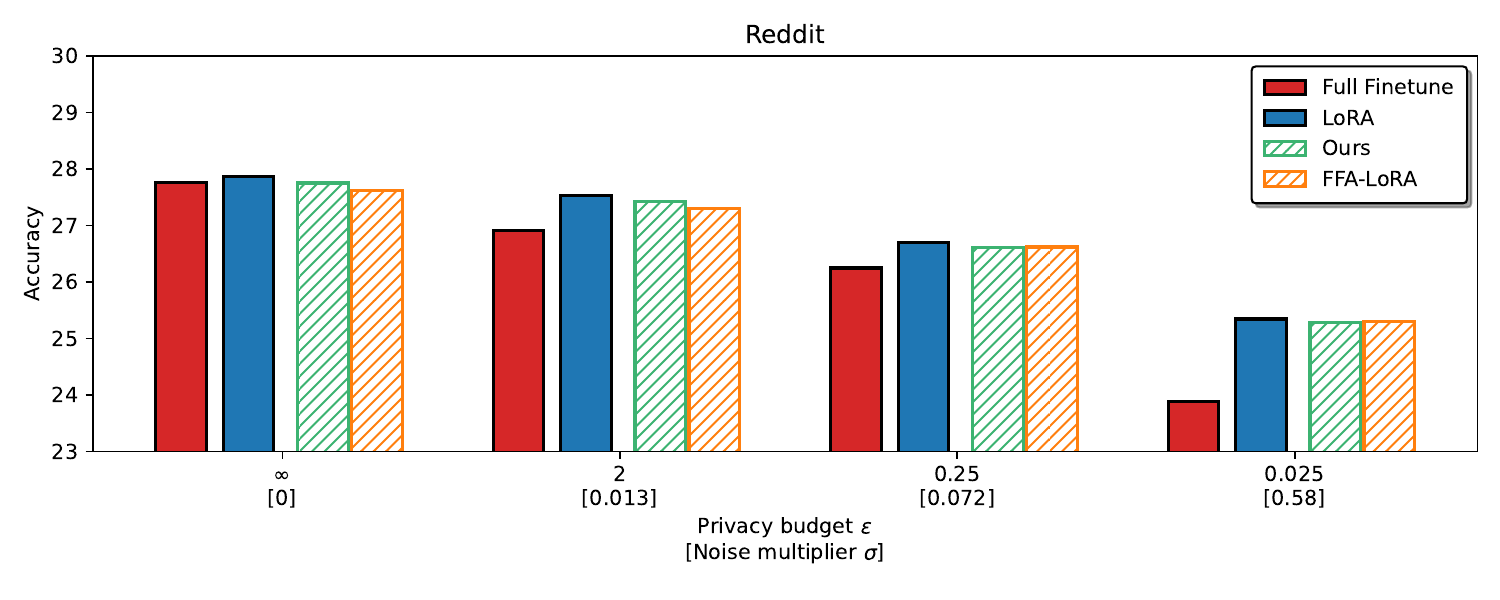}~    \includegraphics[width=0.37\textwidth]{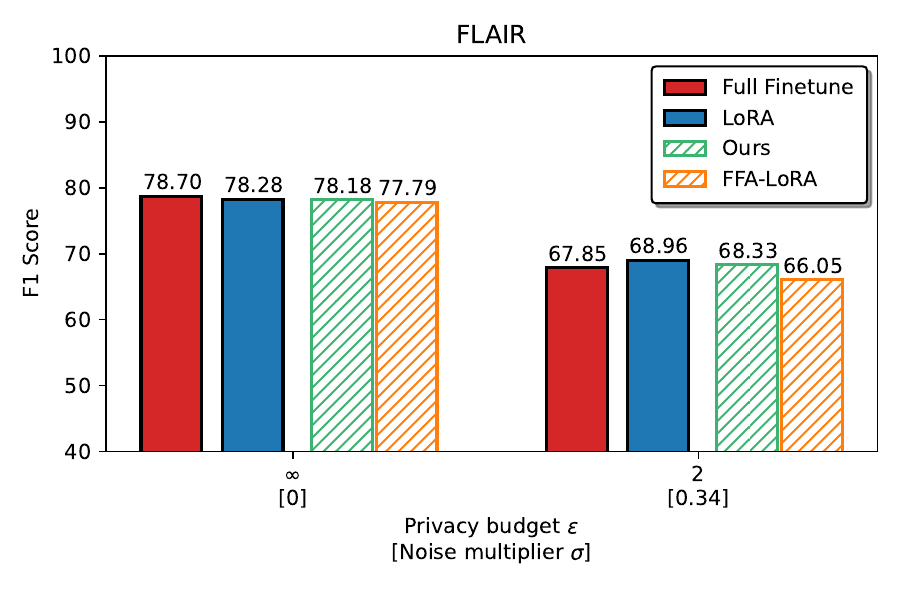}
    \vspace{-0.4cm}
    \caption{We compare the utility $(\uparrow)$ of various finetuning methods optimized with DP-\fedadam. Hatched bars indicate methods which reduce communication by 50\% on top of~\lora. Full finetuning degrades heavily in private settings, while \lora~is relatively more robust. We find that \ffalora~generally sacrifices utility across all settings.}
    \vspace{-0.2cm}
    \label{fig:dp}
\end{figure}

\subsection{Privacy}\label{sec:results_dp}
Finally, while FL provides a base level of privacy by keeping data at the client, FL model updates have been shown to be susceptible to privacy leakage and adversarial attacks~\citep{geiping2020inverting}. To address these concerns, existing work has turned to differential privacy (DP), a popular framework that adds randomness to an algorithm in order to mask example-level contributions that affect the algorithm's output~\citep{mcmahan2017learning}. This provides a probabilistic guarantee that an adversary cannot determine whether a given example was used to generate the output. While early work in centralized settings has shown that \lora~is much more effective than full finetuning at handling DP noise~\citep{luo2021scalable,xu2024dp}, a recent work called \ffalora~suggests that \lora~can amplify noise from DP methods as it decomposes the trainable weights into a product of two low-rank matrices. To address this, \ffalora~(Federated Freeze A)~freezes the $A$ matrix in \lora~and only trains $B$. 

\ffalora~is designed for the stronger notion of \textit{local} (as opposed to \textit{global}) differential privacy. In local DP, clients locally run DP-SGD, which is
only feasible in \textit{cross-silo} FL settings where each client has a large number of local examples. In contrast, we focus on cross-device settings which assume a large widespread pool of clients with few examples per client. To apply global DP to \fedadam, clients upload updates computed by non-private SGD. The server clips these updates, aggregates them, normalizes by the clipping norm, and then adds Gaussian noise with scale $\sigma$~\citep{de2022unlocking}. This protects client privacy at a coarse-grained level where the ``neighboring datasets'' definition of DP applies to the addition or removal of one client's local dataset rather than a single example~\citep{mcmahan2017learning}.

Next, we compare full finetuning, \lora~$(r=16)$, \ours, and \ffalora~in private FL settings. We test four levels of privacy on Reddit and two for FLAIR. We sample a relatively small number of clients (10 for Reddit, 200 for FLAIR) while reporting $\varepsilon$ values with a larger simulated cohort size (1000 for Reddit, 5000 for FLAIR)~\citep{song2022flair}.

\begin{wrapfigure}{h}{0.5\textwidth} 
\includegraphics[width=0.5\textwidth]{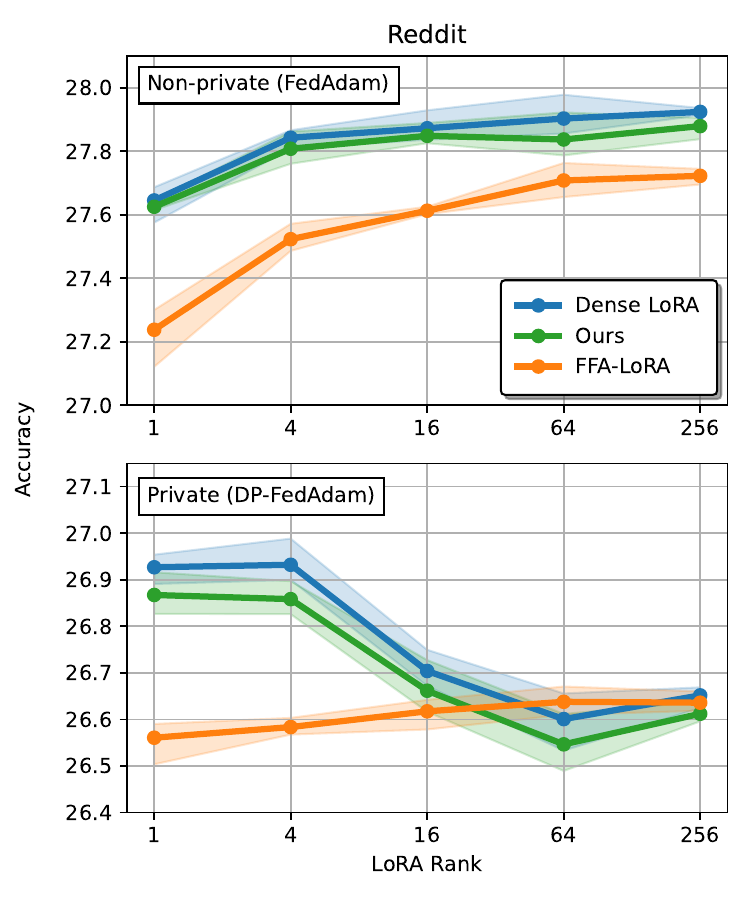}
\vspace{-0.3in}
\caption{We use DP-FedAdam on Reddit to train LoRA modules of various ranks while reducing communication by $\sim 50\%$ with sparsity-based methods.}
\vspace{-0.2in}
\label{fig:dp_params}
\end{wrapfigure}

In Figure~\ref{fig:dp}, we find that privacy degrades the utility of full finetuning much more than \lora-based methods. Next, we find that \ffalora~fails to improve over \lora~in both private and non-private settings, but can still outperform full finetuning. Although the limited performance of \ffalora~is surprising, we believe that this finding is reasonable, as we broadly observe that freezing degrades utility relative to full finetuning. To better understand how privacy can affect this tradeoff, we conduct another experiment that varies the rank of the \lora~module.

Private full finetuning leads to utility loss since the relative scale of the noise becomes much larger than the per-coordinate signal in the update vector~\citep{luo2021scalable}. To investigate this effect, we train \lora~modules with rank from $1$ to $256$. In Figure~\ref{fig:dp_params}, we find that a larger rank does better in non-private settings and vice versa for smaller ranks. Next, we halve communication using \ours~or \ffalora. Overall, we find that \lora~and \ours~achieve higher accuracy in both settings. While we find that freezing can be beneficial in private settings (e.g. fixing $r=64$), \ours~can achieve higher accuracy while using a much lower rank.

%% file: src/conclusion.tex
In this paper, we introduce~\ours, an efficient FL method that significantly reduces the communication cost of \lora. Our method provides both higher utility and substantial communication savings relative to existing pruning-based methods. Our results show that efficient fine-tuning approaches can be made an order of magnitude more efficient when considering FL constraints, highlighting the importance of tailoring efficiency to the setting at hand. Furthermore, we find that \ours~is competitive with specific solutions for other FL concerns of heterogeneity and privacy while achieving superior communication efficiency. Overall, our results indicate that \ours~can serve as a strong baseline for future works in federated fine-tuning.

Still, many important questions remain on how to make \lora~even more efficient in FL.  In order to make ``communication-efficient'' methods truly useful, future work should consider broader bottlenecks in FL such as local training time and client availability. Another important question is whether \lora~is sufficient for practical FL use cases and how to scale to even larger models when \lora~is insufficient. In the future, we aim to investigate such questions and design methods to make high-quality models more accessible to low-resource users.

%% file: src/appendix.tex
\section{Methods}\label{appendix:methods}
\input{src/methods}

\section{Experiment Details}
\subsection{Artifacts}
We provide the code for our experiments on Github: 
\href{https://github.com/imkevinkuo/flasc}{https://github.com/imkevinkuo/flasc}. Setup~instructions  can be found in the README.md file.

\subsection{Compute Resources}
We simulate FL training on a single Nvidia GeForce 1080 Ti GPU (12 GB memory) and parallelize trials over multiple GPUs.

\subsection{Hyperparameter Space}
\label{appendix:hps}
\fedadam~hyperparameters:
\begin{itemize}
    \item Learning rates (Figures~1-6): $\eta_{\text{server}},\eta_{\text{client}} \in [10^{-4}, 5*10^{-4}, 10^{-3}, 5*10^{-3}]$
    \item Learning rates (Figures~7-8):     $\eta_{\text{server}} \in [10^{-3}, 2*10^{-3}, 5*10^{-3}, 10^{-2}, 2*10^{-2}]$
    \item Betas: $\beta_1 = 0.9, \beta_2=0.999$
    \item Clients per round: 10 (CIFAR10, 20Newsgroups, Reddit), 200 (FLAIR)
    \item Client optimizer: SGD (batch size$=16$, momentum$=0.9$)
\end{itemize}

Figure~\ref{fig:methods} (all datasets):
\begin{itemize}[leftmargin=15pt]
    \item \lora~rank: $r \in [1,4,\textbf{16},64]$
    \item \adapterlth~density: $p_{\text{LTH}} \in [0.97, \textbf{0.98}, 0.99]$
    \item \sparseadapter~density: $p \in [1/16, \textbf{1/4}]$
    \item \ours~density: $p_\text{down}=p_\text{up} \in [1/16, \textbf{1/4}]$
    \item Label heterogeneity (CIFAR10 and 20NewsGroups): $\alpha = 0.1$
\end{itemize}
Figure~\ref{fig:time} (20NewsGroups):
\begin{itemize}
    \item \adapterlth~density: $p_{\text{LTH}} = 0.98$
    \item \sparseadapter~density: $p = 1/4$
    \item \ours~density: $p_\text{down} = 1/4, p_\text{up} \in [1/64, 1/16, 1/4]$
\end{itemize}

Figure~\ref{fig:iid_freezing} (CIFAR10):
\begin{itemize}
    \item \lora~Rank: $r = 16$
    \item Density: $d \in [1, 1/4, 1/16, 1/64, 1/256]$
\end{itemize}

Figure~\ref{fig:iid} (CIFAR10, 20NewsGroups):
\begin{itemize}
    \item \lora~rank: $r \in [1,4,16]$
    \item \ours~density: $p_\text{down}=p_\text{up} \in [1/16, \textbf{1/4}]$
    \item Label heterogeneity: $\alpha \in [10^{2}, 1, 10^{-2}]$
\end{itemize}

Figure~\ref{fig:dp} (FLAIR):
\begin{itemize}
    \item Client learning rate $\eta_{\text{client}} = 10^{-2}$
    \item Noise multiplier: $\sigma \in [0, 0.34]$
    \item Clipping norm: $C = 5*10^{-3}$
\end{itemize}

Figure~\ref{fig:dp},\ref{fig:dp_params} (Reddit):
\begin{itemize}
    \item Client learning rate $\eta_{\text{client}} = 5*10^{-4}$
    \item Noise multiplier: $\sigma \in [0, 0.013, 0.072, 0.58]$
    \item Clipping norm: $C = 10^{-4}$
\end{itemize}

\subsection{Simulating privacy noise}
To obtain strong privacy guarantees when using DP-\fedadam, we must bound the sensitivity of the aggregate update with respect to any individual client. The most obvious way to achieve this is to sample a large cohort of clients~\citep{charles2021large}. However, when running experiments with private FL, this can make training costs prohibitively expensive. To make simulation feasible in terms of wall-clock time, a common trick is to select a large (`simulated') client cohort size, compute the noise scale according to the privacy constraints, and then linearly scale it down according to a smaller cohort size actually used for experiments. For instance, \citet{song2022flair} (Sec.~5.1, p.7) uses ``200 users sampled per round to simulate the noise-level with a cohort size of 5,000''. We follow this simulation setup for our experiments on FLAIR. For Reddit, we sample 10 users per round and simulate the noise-level with a cohort size of 1,000. Note that the simulated cohort size only affects the final privacy budget we report and does not change model training or utility.

\pagebreak


%% file: src/methods.tex
\fedadam~is an FL optimization method that accelerates convergence by manipulating the aggregated update at the server~\citep{reddi2020adaptive}. At every round, each participating client $i$ will download a copy of the global weights $W$, fine-tune $W$ to obtain updated weights $W_i'$, and upload $\Delta W_i = W - W_i'$ to the server. The server then computes an average update $\Delta W = \frac{1}{n}\sum_{i=1}^n \Delta W_i$, where $n$ is the number of clients sampled per round. The average may optionally be weighted by each client's dataset size. $\Delta W$ can be interpreted as a global pseudo-gradient; for example, the update rule for FedAvg is to set $W \gets W - \Delta W$ for the next round~\citep{mcmahan2017communication}. In the case of FedAdam, the server maintains a stateful Adam optimizer that takes $\Delta W$ as input and outputs an adapted global update at each round.

\lora~is a reparameterization-based PET method that updates a weight matrix $W \in \mathbb{R}^{d \times k}$ in a low-rank subspace. \lora~freezes $W$ and defines the update $\Delta W \in \mathbb{R}^{d \times k}$ as a product $BA$ where $B \in \mathbb{R}^{d\times r}$ and $A\in\mathbb{R}^{r\times k}$ are newly inserted trainable parameters. By selecting $r$ to be a small constant, $A$ and $B$ can have much fewer entries than $W$. To apply \lora~to FL, we simply treat the adapter weights $A,B$ as the global trainable weights for \fedadam.

\textbf{Pruning} methods rank parameters by magnitude and \textit{prune} the lowest-ranked fraction of parameters, \textit{setting them to zero and freezing them for the rest of training}. To apply pruning to \lora, we prune entries in the adapters $A$ and $B$ while leaving the pretrained weights $W$ intact. In the context of FL, pruning is applied \textit{globally}; the sparsity structure is determined using the global weights at the server and all clients fine-tune the same sparse set of weights. Under this scheme, pruning naturally reduces communication costs as \textit{clients do not have to upload or download zeroed-and-frozen weights}.

\adapterlth~(Lottery Ticket Hypothesis) is a centralized method that alternates between pruning away a small fraction of the lowest magnitude weights and retraining the remaining weights of an adapter module such as \lora~\citep{frankle2018lottery,wu2022pruning}. To use this method in FL, the server prunes the model after every few FL rounds. We use the efficient ``fine-tuning'' version of LTH which continues training from the pruned state rather than rewinding the weights after pruning~\citep{renda2019comparing}. This allows the model to recover from pruning within fewer rounds and is necessary to keep communication costs competitive with the dense \lora~baseline.

\sparseadapter~generally proposes pruning adapters once at initialization~\citep{wu2022pruning,he2022sparseadapter}. For the choice of parameter scoring function, SNIP (gradient-magnitude product) was found to work the best among other baselines~\citep{lee2018snip}. However, magnitude-based scoring does not directly extend to \lora. Because the $B$ matrix in \lora~is initialized to all zeros, magnitude pruning below $40\sim 60\%$ density (depending on the relative input/output layer sizes where \lora~is inserted) would remove all of the $B$ weights and prevent the \lora~modules from training. To fairly evaluate pruning-at-initialization methods, we perform an initial round of FL to train the dense \lora~weights, apply magnitude pruning to the aggregated weights, then train the remaining sparse weights normally.